\documentclass[letterpaper,10pt,conference]{ieeeconf}
\IEEEoverridecommandlockouts
\usepackage[T1]{fontenc}
\usepackage{cite}
\usepackage{amsmath,amssymb,amsfonts}
\usepackage[ruled,vlined,linesnumbered]{algorithm2e}
\usepackage{booktabs}
\usepackage{graphicx}
\usepackage{placeins}
\graphicspath{{./}}
\usepackage{multirow}
\usepackage{xcolor}
\usepackage{caption}
\usepackage{tikz}
\usetikzlibrary{arrows.meta,backgrounds,calc,positioning}
\makeatletter
\let\NAT@parse\undefined
\makeatother
\usepackage{hyperref}
\hypersetup{
    colorlinks=true,
    linkcolor=black,
    citecolor=blue,
    urlcolor=blue
}
\usepackage{adjustbox}
\def\BibTeX{{\rm B\kern-.05em{\sc i\kern-.025em b}\kern-.08em T\kern-.1667em\lower.7ex\hbox{E}\kern-.125emX}}
\usepackage{siunitx}
\usepackage{cleveref}
\begin{document}
\bstctlcite{BSTcontrol}

\title{\LARGE\bfseries
AutoIntervene: Calibrated Intervention\\ for Action-Chunking Imitation Learning Policies
}
\author{Jinhe Tang$^{1}$ and Weiming Zhi$^{1,2,3,*}$\\
{\small $^{1}$School of Computer Science and $^{2}$Australian Center For Robotics, The University of Sydney, Australia}\\
{\small $^{3}$College of Connected Computing, Vanderbilt University, TN, USA}\\
{\small $^{*}$Corresponding author: \texttt{Weiming.Zhi@sydney.edu.au}}\\
{\small Project website: \href{https://aus.bot/research/autointervene/}{\texttt{https://aus.bot/research/autointervene/}}}}
\maketitle
\vspace*{-1.2em}
\thispagestyle{empty}
\pagestyle{empty}

\begin{abstract}
Action-chunking visuomotor policies learn from demonstrations and improve temporal consistency by predicting short action sequences rather than single-step commands. Yet perception errors and execution drift can move the robot outside the demonstration distribution, while the policy continues to produce smooth action chunks that are inconsistent with the observed state. We present \emph{AutoIntervene}, an online framework that selectively transfers control between an action-chunking policy and an operator during deployment. \emph{AutoIntervene} evaluates proposed chunks against a visual-action support memory built from successful task executions, combining visual similarity with consistency between proposed and reference actions. Phase-local support governs policy-to-operator transfer within the current task phase, whereas global support governs the return to policy control after operator recovery. We calibrate separate switching thresholds for the two directions from empirical quantiles of evaluation-level scores on held-out expert demonstrations, avoiding direct manual tuning of score cutoffs. Intervention segments retained from successful rollouts target learner-induced states and provide corrective supervision for subsequent policy updates. Experiments on real-world bimanual manipulation tasks show higher post-adaptation task success and lower operator-control time than manual intervention. Videos and additional results are available at \url{https://aus.bot/research/autointervene/}.
\end{abstract}

\section{Introduction}
\label{sec:introduction}
Imitation learning provides a practical framework for learning robot policies from demonstrations~\cite{Osa_2018}, including stable motion generators that adapt to environmental changes~\cite{zhi2022diffeomorphic}, while offline human data has enabled complex multi-stage and long-horizon manipulation~\cite{pmlr-v164-mandlekar22a,mandlekar2020gti}. Recent action-chunking policies improve temporal consistency by predicting short sequences of future actions rather than single-step commands, enabling visuomotor policies to execute contact-rich manipulation tasks without hand-designed controllers~\cite{zhao2023learningfinegrainedbimanualmanipulation}. However, these policies remain brittle under deployment-time distribution shift~\cite{ross2011reductionimitationlearningstructured}. DART broadens demonstration coverage through noise injection but provides no deployment-time mechanism for detecting unsupported policy behaviour or requesting corrective control~\cite{laskey2017dartnoiseinjectionrobust}. Small perception errors, missed contacts, or accumulated execution error can move the robot into states that are poorly covered by the demonstration data. Once this occurs, an action-chunking policy can continue producing smooth action chunks while no longer making task progress, leading to failures such as misaligned grasps or incomplete subtask transitions~\cite{agia2024unpackingfailuremodesgenerative,zheng2026rewindil}.

In this paper, we introduce \emph{AutoIntervene}, an online intervention framework for improving the deployment reliability of action-chunking policies. AutoIntervene evaluates each proposed action chunk against successful visual-action support and transfers control when the proposal repeatedly becomes unsupported. It uses phase-local references to decide when control should pass from the policy to the operator and global references to decide when it should return to the policy, allowing autonomy to resume from any supported task phase after operator correction. At each adaptation round, separate switching thresholds are calibrated offline by recomputing proposals and support scores on held-out successful expert demonstrations under the corresponding retrieval scope.

\begin{figure}
    \centering
    \includegraphics[width=1.0\linewidth]{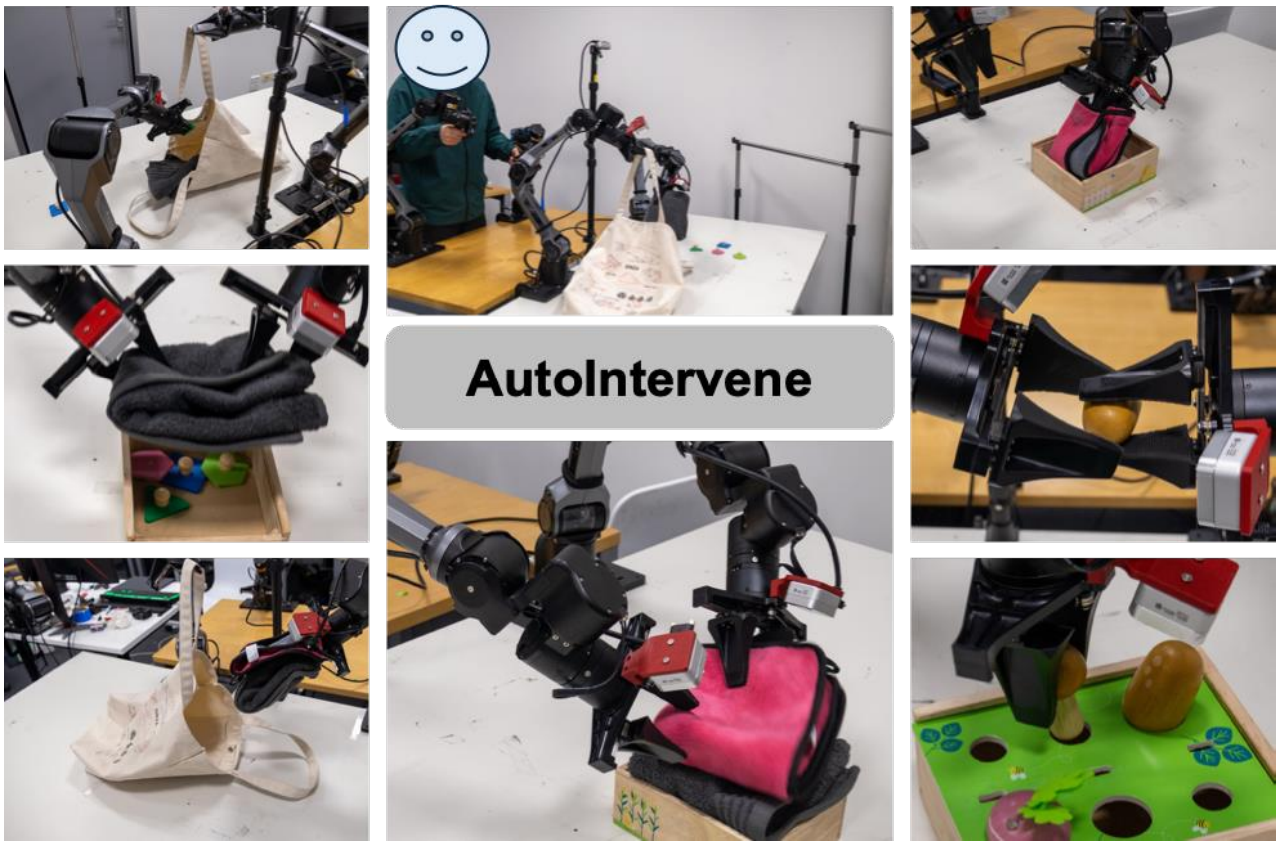}
    \caption{
    AutoIntervene transfers control between the autonomous policy and the operator, thereby guiding the collection of targeted intervention data. We evaluate the system across diverse manipulation tasks.
    }
    \label{fig:overview_executions}
\end{figure}

Operator-controlled segments from successful rollouts are retained for subsequent policy updates, yielding a selective DAgger-style loop~\cite{kelly2019hgdaggerinteractiveimitationlearning,hu2025rac}. Unlike standard DAgger, which queries expert actions at learner-visited states throughout a rollout~\cite{ross2011reductionimitationlearningstructured}, AutoIntervene requests supervision only during automatically identified unsupported periods. This connects deployment-time monitoring with targeted data aggregation. \Cref{fig:overview_executions} illustrates representative executions, including manipulation of deformable materials such as towels and fabric tote bags.

The contributions of this work are threefold:
\begin{enumerate}
    \item We introduce a bidirectional intervention framework for action-chunking imitation policies, using phase-local support for policy-to-operator transfer and global support for the return to policy control.

    \item We develop a mode-specific visual-action calibration procedure that uses held-out successful expert demonstrations to recompute separate intervention and recovery thresholds at each adaptation round under their respective retrieval scopes.

    \item We evaluate AutoIntervene on nine real-world tasks and show that targeted intervention trajectories support iterative policy improvement with substantially less operator-control time than collecting additional full demonstrations.
\end{enumerate}

\section{Related Work}

\textbf{Visuomotor imitation policies.}
Visuomotor imitation policies differ in how they represent temporally extended actions. Action chunking predicts short action sequences directly, whereas diffusion and flow-matching policies generate them through iterative or continuous-time processes~\cite{zhao2023learningfinegrainedbimanualmanipulation,Chi-RSS-23,zhang2025flowpolicy}. Streaming flow trajectories can also be constrained post-training~\cite{long2026safepolicies}. Although these approaches improve action generation, they do not by themselves determine when a deployed proposal has left demonstrated support or when control should return after intervention. AutoIntervene addresses this deployment layer independently of the action head. We evaluate it with Action Chunking with Transformers (ACT), Diffusion Policy, and Flow Matching heads.

\textbf{Interactive imitation learning.}
Interactive policy adaptation traditionally relies on an operator to monitor policy execution and manually decide when to take over and return control. Robot-gated methods instead request intervention automatically from deployment-time signals. LazyDAgger uses policy--expert action discrepancy to trigger expert involvement, whereas RND-DAgger uses state novelty estimated by random network distillation~\cite{hoque2021lazydaggerreducingcontextswitching,bire2024efficientactiveimitationlearning}. AutoIntervene follows this robot-gated direction but treats control transfer as a bidirectional, support-based process: phase-local support governs policy-to-operator transfer, global support governs operator-to-policy return, and retained recovery segments provide corrective data for subsequent policy adaptation.

\textbf{Runtime policy monitoring.}
Runtime monitors can be compared by the signal they inspect and the action taken after detection. Error-Aware Imitation Learning detected potential failures from teleoperation data~\cite{pmlr-v164-wong22a} and ConditionNET learned action-conditioned preconditions and effects~\cite{sliwowski2025conditionnet}. Sentinel combined temporal action inconsistency with vision-language progress checks~\cite{agia2024unpackingfailuremodesgenerative}, whereas FAIL-Detect estimated failure uncertainty from successful data alone~\cite{xu2025detectfailuresfailuredata}. Trajectory-level out-of-distribution detection has also been formulated directly on $\mathrm{SE}(3)$ pose sequences using diffusion models~\cite{cheng2025dose3}. PATCH conditions localized latent-patch innovation on the active action chunk to pause and resume policy execution under local scene disturbances~\cite{zhou2026patch}. Rewind-IL further coupled calibrated inter-chunk discrepancy with respawning at a semantically verified safe state~\cite{zheng2026rewindil}. These approaches respectively target trajectory-level out-of-distribution detection, localized transient disturbances, or checkpoint-based respawning. AutoIntervene instead uses retrieval-based visual-action support to govern both policy-to-operator and operator-to-policy transfer, then converts operator-controlled recovery into corrective training data.

\section{AutoIntervene}
\label{sec:autointervene}

\begin{figure}[t]
    \centering
    \includegraphics[width=\columnwidth]{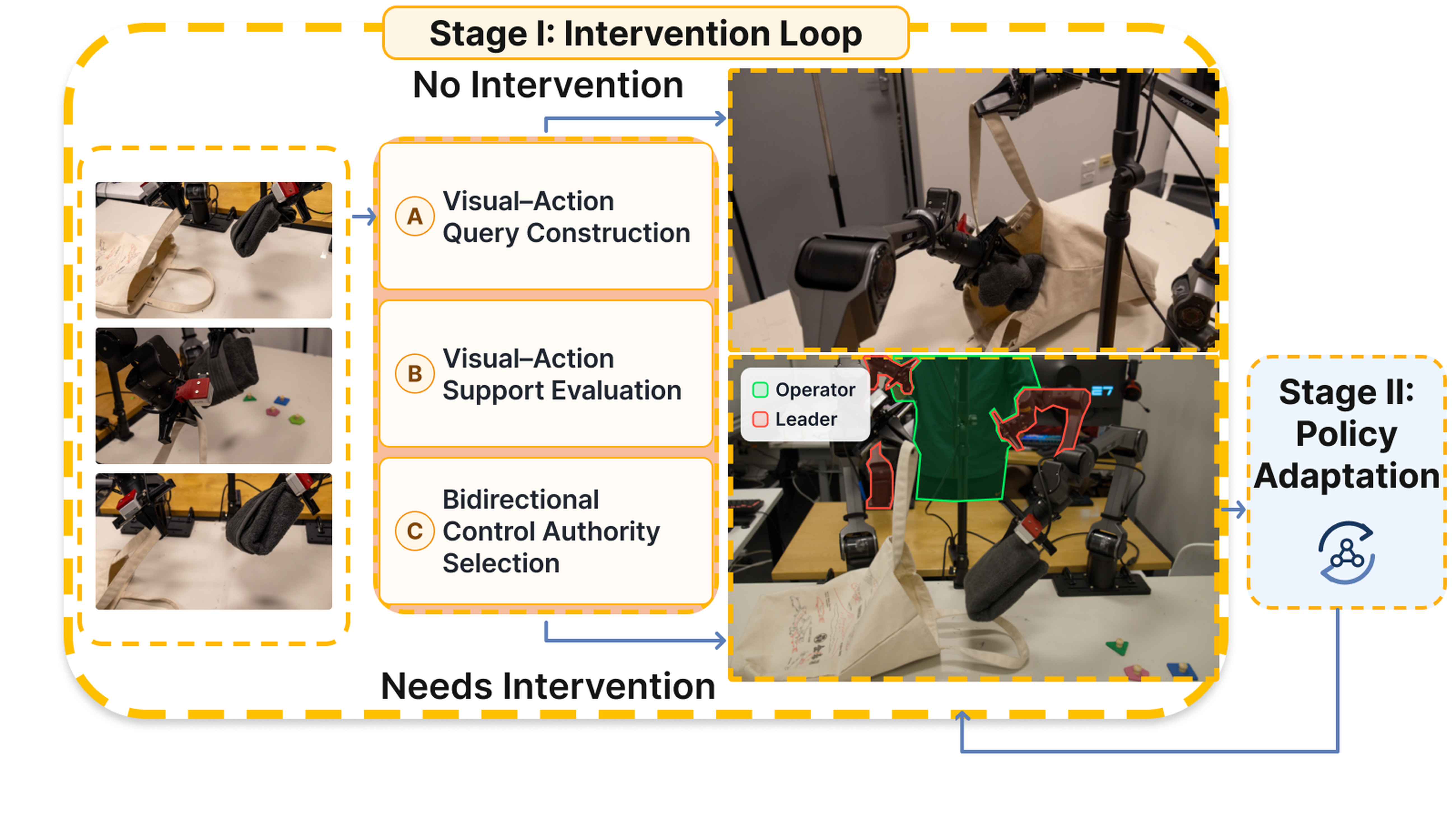}
    \caption{Overview of \emph{AutoIntervene}. During Stage I, the intervention loop sequentially (A) constructs a visual-action query from the current visual observations and policy-proposed action chunk, (B) evaluates visual-support and action-risk scores, and (C) selects control authority using calibrated acceptance criteria. The retained intervention segments are subsequently used to adapt the policy in Stage II, after which the updated policy is redeployed.}
    \label{fig:overview}
\end{figure}

AutoIntervene is designed to improve a deployed action-chunking visuomotor policy by converting automatically triggered operator interventions into targeted supervision. As illustrated in \Cref{fig:overview}, it consists of two stages. The first stage, \textbf{Intervention Loop}, monitors the deployed policy, transfers control to the operator when intervention is required, returns control to the policy when its proposed actions regain sufficient support, and records the resulting intervention segments. The loop proceeds sequentially through three modules: \textbf{(A) Visual-Action Query Construction}, \textbf{(B) Visual-Action Support Evaluation}, and \textbf{(C) Bidirectional Control Authority Selection}. Together, these modules construct a visual-action query from the current visual observations and policy-proposed action chunk, map it to visual-support and action-risk scores, and select control authority using their calibrated acceptance criteria. The second stage, \textbf{Policy Adaptation}, uses the retained intervention segments to update the current policy. The updated policy is subsequently redeployed, and the intervention loop collects new intervention segments for the next adaptation round.

\vspace{0.3em}\noindent\textbf{Control setup.}
We implement the intervention loop using ALOHA-style leader--follower teleoperation with TriPilot-FF-style arm-side force reflection~\cite{zhao2023learningfinegrainedbimanualmanipulation,li2026tripilotff}. The leader arms serve as the operator input, while the follower arms interact with the task. Under policy control, the leader and follower arms receive the same policy-generated joint-and-gripper commands and remain aligned. When control passes to the operator, each follower arm switches to tracking the corresponding operator-manipulated leader arm, allowing correction to begin without repositioning the leader arms.

\subsection{Visual-Action Query Construction}

During deployment, visually similar observations may recur at task phases that require different actions~\cite{sermanet2018tcn,ouyang2025scil,li2026trace}. Visual similarity alone is therefore insufficient to assess whether the policy's proposed motion is appropriate. This motivates AutoIntervene to evaluate the current observation jointly with the action chunk that the policy is about to execute. At evaluation time $t$, the current policy $\pi$ takes the current camera images as input. Its visual encoder maps these images to embeddings $E_t=(e_t^{(1)},\ldots,e_t^{(C)})$, where $C$ denotes the number of camera views. The policy's action head predicts an $H$-step action chunk, and the monitor uses its first $H_r$ steps for evaluation, where $1\leq H_r\leq H$. We denote this predicted prefix by $A_t$.

For our bimanual system, we group the commands in $A_t$ by arm: one group contains the left-arm joint-and-gripper commands, and the other contains the corresponding right-arm commands. Together, the visual embeddings and predicted action prefix form the visual-action query $\mathcal Q_t=(E_t,A_t)$.

\subsection{Visual-Action Support Evaluation}

Given the visual-action query $\mathcal Q_t$, AutoIntervene uses a visual-action
memory $\mathcal M$ constructed from behaviour references and
summarises their high-dimensional comparison into two scalar quantities: a
visual-support score and an action-risk score.

\vspace{0.3em}\noindent\textbf{Visual-action memory and mode-specific retrieval.}
The visual-action memory is constructed from all trajectories used to train the
current policy $\pi$. Suppose that there are $N$ such trajectories, collected
in $\mathcal D=\{\tau_i\}_{i=1}^{N}$, where $\tau_i$ is the $i$-th trajectory
in the collection. Each trajectory in $\mathcal D$ is used to generate the
corresponding visual-action memory entries.

\begin{figure}[t]
    \centering
    \includegraphics[width=\columnwidth]{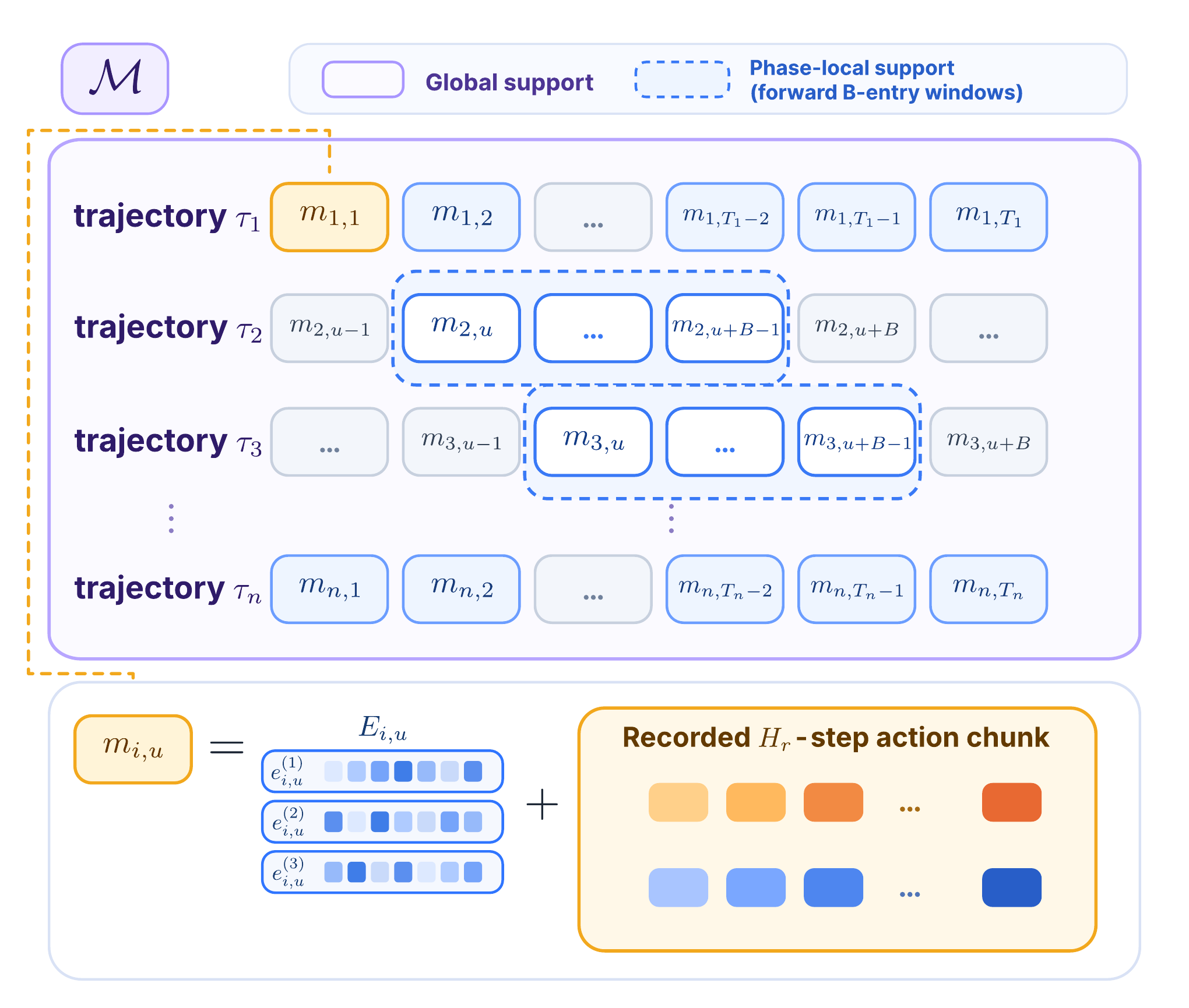}
    \caption{Memory construction and mode-specific retrieval. $\mathcal R_{\mathrm{op},t}$ uses all entries, whereas $\mathcal R_{\mathrm{pol},t}$ combines forward $B$-entry windows from $J$ selected trajectories.}
    \label{fig:visual_action_memory}
\end{figure}

For any trajectory $\tau_i\in\mathcal D$, its length is $|\tau_i|$ recorded
timesteps, and it contributes $|\tau_i|-H_r+1$ memory entries. These entries
correspond to all starting timesteps from which a complete $H_r$-step action
chunk can be extracted. For any such entry, let
$u\in\{1,\ldots,|\tau_i|-H_r+1\}$ be its starting timestep. The $H_r$
recorded actions from timestep $u$ through timestep $u+H_r-1$ form the action
chunk $A_{i,u}=(a_{i,u},\ldots,a_{i,u+H_r-1})$, while $E_{i,u}$ is the
multi-view visual embedding of trajectory $\tau_i$ at the same starting
timestep. Together, $A_{i,u}$ and $E_{i,u}$ form the memory entry
$m_{i,u}=(E_{i,u},A_{i,u})$. Collecting the entries generated from all valid
starting timesteps across all trajectories yields the complete visual-action
memory $\mathcal M$, as illustrated in Figure~\ref{fig:visual_action_memory}.

Once the visual-action memory $\mathcal M$ has been constructed, AutoIntervene
uses it at a mode-specific evaluation frequency during deployment to continually assess
the current visual-action query. At each evaluation time $t$, either the policy
or the operator controls the bimanual system, so AutoIntervene determines the
memory entries used for the current assessment according to the active control mode.
Let $\beta\in\{\mathrm{pol},\mathrm{op}\}$ denote this mode, where
$\beta=\mathrm{pol}$ indicates that the policy controls the bimanual system and
$\beta=\mathrm{op}$ indicates that the operator controls it. The corresponding retrieval set is denoted by
$\mathcal R_{\beta,t}\subseteq\mathcal M$.

When $\beta=\mathrm{op}$, the operator controls the bimanual system while the
policy continues to generate predictions in the background. The operator's
recovery may change the object state or complete part of the task, so the
current state need not remain near the task phase at which the intervention
began. To allow the policy to regain support from any valid task phase reached
after recovery, AutoIntervene retrieves from the complete visual-action memory.
We refer to support evaluated under this complete-memory retrieval scope as
global support.

When $\beta=\mathrm{pol}$, the policy controls the bimanual system and task
progress is locally continuous across successive evaluations. However, memory
entries from different task phases may contain similar visual embeddings but
different recorded action chunks. Repeatedly searching the complete memory
could therefore introduce a match from the wrong phase~\cite{ouyang2025scil}.
Whenever the policy begins controlling the bimanual system, its first
evaluation compares the current visual embedding $E_t$ with the visual
embedding $E_{i,u}$ contained in every memory entry
$m_{i,u}\in\mathcal M$. AutoIntervene retains the best-matching memory entry on
each training trajectory $\tau_i$ and then selects the $J$ trajectories whose
retained entries have the strongest matches. Each selected trajectory
contributes a forward window containing its matched entry and at most the next
$B-1$ entries. Subsequent evaluations retrieve only within these windows. After
every $U_{\mathrm{win}}$ executed policy actions, each window
advances along its trajectory according to the newly matched entry without
returning to an earlier task phase. Let
$\mathcal W_{j,t}$ denote the window from the $j$-th selected trajectory at
evaluation time $t$, where $j=1,\ldots,J$. Thus, the mode-specific retrieval
set is defined as below. We refer to support evaluated within these forward
windows as phase-local support.
\[
    \mathcal R_{\beta,t}
    =
    \begin{cases}
        \mathcal M,
        & \beta=\mathrm{op},\\[2pt]
        \mathcal W_{1,t}\cup\cdots\cup\mathcal W_{J,t},
        & \beta=\mathrm{pol}.
    \end{cases}
\]

For notational simplicity, throughout visual-neighbour retrieval and
action-risk scoring, we write a generic memory entry as $m=(E,A)$ and
suppress its trajectory and timestep indices.

\vspace{0.3em}\noindent\textbf{Visual-neighbour retrieval.}
For a reference entry $m\in\mathcal R_{\beta,t}$, we take its multi-view visual
component $E=(e^{(1)},\ldots,e^{(C)})$ and compare it view by view with the
current query embedding $E_t=(e_t^{(1)},\ldots,e_t^{(C)})$. Their minimum
cross-view visual similarity is
\begin{align}
    s_t(m)
    &=
    \min_{1\leq c\leq C}
    \operatorname{sim}\!\left(e_t^{(c)},e^{(c)}\right),
    \label{eq:min_camera_similarity}
\end{align}
In \eqref{eq:min_camera_similarity}, $\operatorname{sim}$ is cosine similarity between $\ell_2$-normalised
embeddings. Taking the minimum prevents a high similarity in one view from
masking a mismatch in another, so every view must support the match. From
$\mathcal R_{\beta,t}$, AutoIntervene retains up to $K$ entries with the largest
similarities as $\mathcal N_{\beta,t}$.

\vspace{0.3em}\noindent\textbf{Action-risk scoring.}
Given $\mathcal N_{\beta,t}$, we compare the proposed action chunk $A_t$ with
the stored action component $A$ of each entry $m\in\mathcal N_{\beta,t}$. The proposal
is the chunk awaiting execution under policy control and the background policy
prediction under operator control. Following the action grouping defined above,
the normalised distance between these chunks for each group $g$ is given below,
where the superscript $(g)$ selects that group's action dimensions:
\begin{align}
    d_{t,g}(m)
    &=
    \left\|
    \frac{A_t^{(g)}-A^{(g)}}{\sigma_g}
    \right\|_{2},
    \label{eq:action_distance}
\end{align}
In \eqref{eq:action_distance}, $\sigma_g$ contains the per-dimension standard deviations of group $g$'s
recorded actions in $\mathcal M$, and division is applied element-wise. The Euclidean distance is computed over
all retained steps and action dimensions.

For each action group $g$, let the
action-selected set $\mathcal P_g$ contain the $M$ entries in
$\mathcal N_{\beta,t}$ with the smallest distances. Aggregating these references
reduces sensitivity to a accidental match while limiting interference
from less action-relevant visual neighbours. Let $g^{\star}$ be the action group
whose selected entries have the largest mean distance, so that risk is governed
by the least-supported group rather than masked by better-matched groups. Using
the same references, the mode-dependent action risk and visual support are
\begin{align}
    r_{\beta,t}
    &=\frac{1}{M}\sum_{m\in\mathcal P_{g^{\star}}}
    d_{t,g^{\star}}(m), \nonumber\\[-0.2em]
    s_{\beta,t}
    &=\frac{1}{M}\sum_{m\in\mathcal P_{g^{\star}}}s_t(m).
    \label{eq:visual_action_aggregation}
\end{align}
As defined in \eqref{eq:visual_action_aggregation}, both scores use the same action-selected references.

Finally, $\bar r_{\beta,t}$ averages up to the $W$ most recent action-risk
values in the current control interval, using all available values before $W$
evaluations. Visual support remains instantaneous so that earlier matches do
not mask a sudden loss of visual correspondence.
The stage returns the score pair $\bigl(s_{\beta,t},\bar r_{\beta,t}\bigr)$ and,
under policy control, the selected references $\mathcal P_{g^{\star}}$ for
retrieval-window updates.

\vspace{0.3em}\noindent\textbf{Policy-side retrieval-window update.}
After every $U_{\mathrm{win}}$ executed policy actions, the
selected references $\mathcal P_{g^{\star}}$ update the phase-local retrieval
windows. For the $j$-th selected trajectory $\tau_{i_j}$, let $u_{j,t}$ denote
its current window position. We update this position to the largest value among
its current position and the starting-timestep indices of the selected
references from the same trajectory:
\begin{align}
    u_{j,t+1}
    &=
    \max\!\left(
    \{u_{j,t}\}\cup
    \{u\mid m_{i_j,u}\in\mathcal P_{g^{\star}}\}
    \right).
    \label{eq:window_update}
\end{align}
The update in \eqref{eq:window_update} is nondecreasing, so each trajectory
window moves forward or remains at its current position.

\subsection{Bidirectional Control Authority Selection}

The control-authority selector receives the score pair
$\bigl(s_{\beta,t},\bar r_{\beta,t}\bigr)$ for the current control mode
$\beta\in\{\mathrm{pol},\mathrm{op}\}$ and decides whether to retain or switch
control after evaluation $t$. This decision requires separate visual-support
and action-risk thresholds for the two control modes.

Let $\mathcal D_{\mathrm{cal}}$ be a fixed held-out set of successful expert
trajectories excluded from policy training and the visual-action memory. Before
deploying the policy in each adaptation round, we apply the same support evaluation to
$\mathcal D_{\mathrm{cal}}$ separately under policy and operator control. This
produces the visual-support collection $S_{\mathrm{cal}}^{\beta}$ and
action-risk collection $R_{\mathrm{cal}}^{\beta}$ for mode $\beta$.

Let $\alpha_s^{\beta}$ and $\alpha_r^{\beta}$ denote the prescribed lower- and
upper-tail rates, respectively. With $\widehat Q_p$ denoting the empirical
$p$-quantile, the corresponding thresholds are
\begin{align}
    \theta_s^{\beta}
    =\widehat Q_{\alpha_s^{\beta}}
    \!\left(S_{\mathrm{cal}}^{\beta}\right), &&
    \theta_r^{\beta}
=\widehat Q_{1-\alpha_r^{\beta}}
    \!\left(R_{\mathrm{cal}}^{\beta}\right).
    \label{eq:mode_specific_thresholds}
\end{align}
In \eqref{eq:mode_specific_thresholds}, $\theta_s^{\beta}$ is the lower visual-support threshold and
$\theta_r^{\beta}$ is the upper action-risk threshold. The two modes are
calibrated separately because their retrieval scopes produce different score
distributions.

During deployment, a policy proposal is accepted when
$s_{\beta,t}\geq\theta_s^{\beta}$ and
$\bar r_{\beta,t}\leq\theta_r^{\beta}$, and is rejected otherwise. Under
policy control, a rejection, an empty phase-local reference set, or all phase-local retrieval windows reaching their final valid chunks increments the policy-side rejection counter $c_{\mathrm{pol},t}$. Acceptance resets it. Under operator control,
the policy is evaluated in the background and $c_{\mathrm{op},t}$ counts
consecutive acceptances. A rejected proposal resets it. Given the
required persistence lengths $L_{\mathrm{pol}}$ and $L_{\mathrm{op}}$, the
authority update $\beta^{+}$ is
\begin{align}
    \beta^{+}
    &=
    \begin{cases}
    \mathrm{op},
    & \beta=\mathrm{pol}\ \text{and}\
    c_{\mathrm{pol},t}\geq L_{\mathrm{pol}},\\[0.35em]
    \mathrm{pol},
    & \beta=\mathrm{op}\ \text{and}\
    c_{\mathrm{op},t}\geq L_{\mathrm{op}},\\[0.35em]
    \beta,
    & \text{otherwise}.
    \end{cases}
    \label{eq:bidirectional_authority}
\end{align}
The persistence lengths in \eqref{eq:bidirectional_authority} require the
corresponding decision to remain consistent across successive evaluations,
preventing brief score fluctuations from causing unnecessary control changes.

\subsection{Policy Adaptation}
\label{sec:policy_adaptation}

In adaptation round $k$, the Intervention Loop deploys policy $\pi^{(k)}$, trained on
the accumulated trajectory collection $\mathcal D^{(k)}$. Each interval during
which the operator controls the system is retained as an intervention segment
and stored as a separate intervention trajectory. The segment begins when control switches from the policy to the operator and
ends when control returns to the policy or the episode terminates. The
intervention trajectories retained from successful rollouts form
$\Delta\mathcal D^{(k)}$. Keeping them separate prevents training
examples from crossing a change in control source and focuses supervision on
states that required operator correction~\cite{ross2011reductionimitationlearningstructured,kelly2019hgdaggerinteractiveimitationlearning}.

At the end of round $k$, the new intervention data are added to the current training
set. For any trajectory collection $\mathcal D$, let $p_{\mathcal D}$ denote
uniform sampling over its valid $H$-step training examples. The next training
set and adaptation sampling distribution are
\begin{align}
    \mathcal D^{(k+1)}
    &=
    \mathcal D^{(k)}
    \cup \Delta\mathcal D^{(k)}, \nonumber\\
    p^{(k+1)}
    &=
    \lambda_{\mathrm{mix}}p_{\mathcal D^{(k)}} \nonumber\\[-0.2em]
    &\quad
    +(1-\lambda_{\mathrm{mix}})
    p_{\Delta\mathcal D^{(k)}},
    \quad 0\leq\lambda_{\mathrm{mix}}\leq1,
    \label{eq:policy_adaptation}
\end{align}
The mixture in \eqref{eq:policy_adaptation} retains prior training data while assigning the new intervention data a
fixed sampling weight~\cite{mandlekar2021humanloopteleoperation,rolnick2019experiencereplay}.

The next policy $\pi^{(k+1)}$ is trained by standard behaviour cloning on
examples sampled from $p^{(k+1)}$. The visual-action memory is then rebuilt
from $\mathcal D^{(k+1)}$ using the visual encoder of $\pi^{(k+1)}$. Before
redeployment, the mode-specific thresholds are recalibrated on
$\mathcal D_{\mathrm{cal}}$, after which $\pi^{(k+1)}$ enters the next
Intervention Loop. Repeating this cycle progressively incorporates intervention data
from unsupported task regions encountered during deployment.

\FloatBarrier
\section{Empirical Evaluation}
\label{sec:experiments}
\begin{table}[t]
\centering
\caption{Definitions of the evaluated tasks.}
\label{tab:appendix_task_definitions}
\footnotesize
\setlength{\tabcolsep}{2pt}
\renewcommand{\arraystretch}{1.12}
\begin{adjustbox}{width=0.99\linewidth}
\begin{tabular}{p{0.3\linewidth} p{0.68\linewidth}}
\toprule
\textbf{Task} & \textbf{Description} \\
\midrule

{Peg Disassembly}
& Disassemble a pegged object and place all parts into the box. \\

{Potato Transfer}
& Transfer a wooden potato between hands and place it into the box. \\

{Towel Folding}
& Fold a towel from a fixed initial configuration. \\

{Towel Bagging}
& Fold a towel, pack it and the pegs into a bag, and move the bag to the target side. \\

{Lidded Box Packing}
& Open the box, pack the objects inside, and close the lid. \\

{Plant Sorting}
& Place three wooden plant objects into their matching box slots. \\

{Towel Box Packing}
& Remove pegs, fold the towel, reposition the box, and place the towel inside. \\

{Two-Towel Box Packing}
& Remove pegs, fold two towels, and place both into the box. \\

{Towels-and-Cable Bagging}
& Fold two towels, pick up the cable, and place all three items into the bag. \\

\bottomrule
\end{tabular}
\end{adjustbox}
\end{table}

\newcommand{\pct}[1]{#1\%}

\begin{table*}[t]
\centering
\caption{
Success (\%) and recorded control-data time across seven tasks. Human and AutoIntervene use manually and automatically triggered interventions, respectively. R1 and R2 denote adaptation rounds. $\Delta t$ is cumulative additional operator-control time, while Initial and Additional Full Data report total demonstration time.
}
\label{tab:main_exp}
\setlength{\tabcolsep}{2.6pt}
\scriptsize
\begin{adjustbox}{width=0.95\linewidth}
\begin{tabular}{
@{}l
r S[table-format=4.1]
r S[table-format=3.1]
r S[table-format=3.1]
r S[table-format=3.1]
r S[table-format=3.1]
r S[table-format=4.1]
@{}}
\toprule
\multirow{2}{*}{Task}
& \multicolumn{2}{c}{Initial}
& \multicolumn{2}{c}{Human R1}
& \multicolumn{2}{c}{Human R2}
& \multicolumn{2}{c}{\textsc{AutoIntervene} R1}
& \multicolumn{2}{c}{\textsc{AutoIntervene} R2}
& \multicolumn{2}{c}{Additional Full Data} \\
\cmidrule(lr){2-3}
\cmidrule(lr){4-5}
\cmidrule(lr){6-7}
\cmidrule(lr){8-9}
\cmidrule(lr){10-11}
\cmidrule(lr){12-13}
& \multicolumn{1}{c}{Succ. (\%)}
& \multicolumn{1}{c}{Time (s)}
& \multicolumn{1}{c}{Succ. (\%)}
& \multicolumn{1}{c}{$\Delta t$ (s)}
& \multicolumn{1}{c}{Succ. (\%)}
& \multicolumn{1}{c}{$\Delta t$ (s)}
& \multicolumn{1}{c}{Succ. (\%)}
& \multicolumn{1}{c}{$\Delta t$ (s)}
& \multicolumn{1}{c}{Succ. (\%)}
& \multicolumn{1}{c}{$\Delta t$ (s)}
& \multicolumn{1}{c}{Succ. (\%)}
& \multicolumn{1}{c}{Time (s)} \\
\midrule
Peg Disassembly    & \pct{16}   & 1173.7 & \pct{52}   & 123.6 & \pct{56}   & 286.9 & \pct{64}   & 49.3  & \textbf{\pct{72}}   & 127.1 & \pct{60}   & 1853.0 \\
Potato Transfer    & \pct{44}   & 368.7  & \pct{52}   & 20.7  & \textbf{\pct{80}}   & 74.2  & \pct{48}   & 29.6  & \pct{76}   & 50.2  & \pct{64}   & 547.3  \\
Towel Folding      & \pct{56}   & 766.2  & \pct{60}   & 36.4  & \textbf{\pct{100}}  & 144.8 & \pct{76}   & 61.4  & \pct{96}   & 140.4 & \pct{68}   & 1100.0 \\
Towel Bagging      & \pct{24}   & 1761.1 & \pct{32}   & 130.2 & \pct{36}   & 266.5 & \pct{44}   & 121.6 & \textbf{\pct{60}}   & 171.0 & \pct{40}   & 2536.7 \\
Lidded Box Packing & \pct{8}    & 989.6  & \pct{20}   & 107.5 & \pct{92}   & 246.7 & \pct{52}   & 63.1  & \textbf{\pct{100}}  & 130.2 & \pct{60}   & 1440.0 \\
Plant Sorting      & \pct{24}   & 450.2  & \pct{52}   & 88.8  & \pct{56}   & 157.1 & \pct{52}   & 76.6  & \textbf{\pct{72}}   & 128.9 & \pct{32}   & 693.1  \\
Towel Box Packing  & \pct{44}   & 1332.4 & \pct{52}   & 36.4  & \pct{60}   & 83.3  & \pct{80}   & 61.4  & \textbf{\pct{84}}   & 112.3 & \pct{68}   & 1930.5 \\
\midrule
Average            & \pct{30.9} & 977.4  & \pct{45.7} & 77.7  & \pct{68.6} & 179.9 & \pct{59.4} & 66.1  & \textbf{\pct{80.0}} & 122.9 & \pct{56.0} & 1442.9 \\
\bottomrule
\end{tabular}
\end{adjustbox}
\end{table*}

\begin{figure*}[t]
\centering
\includegraphics[width=0.24\linewidth]{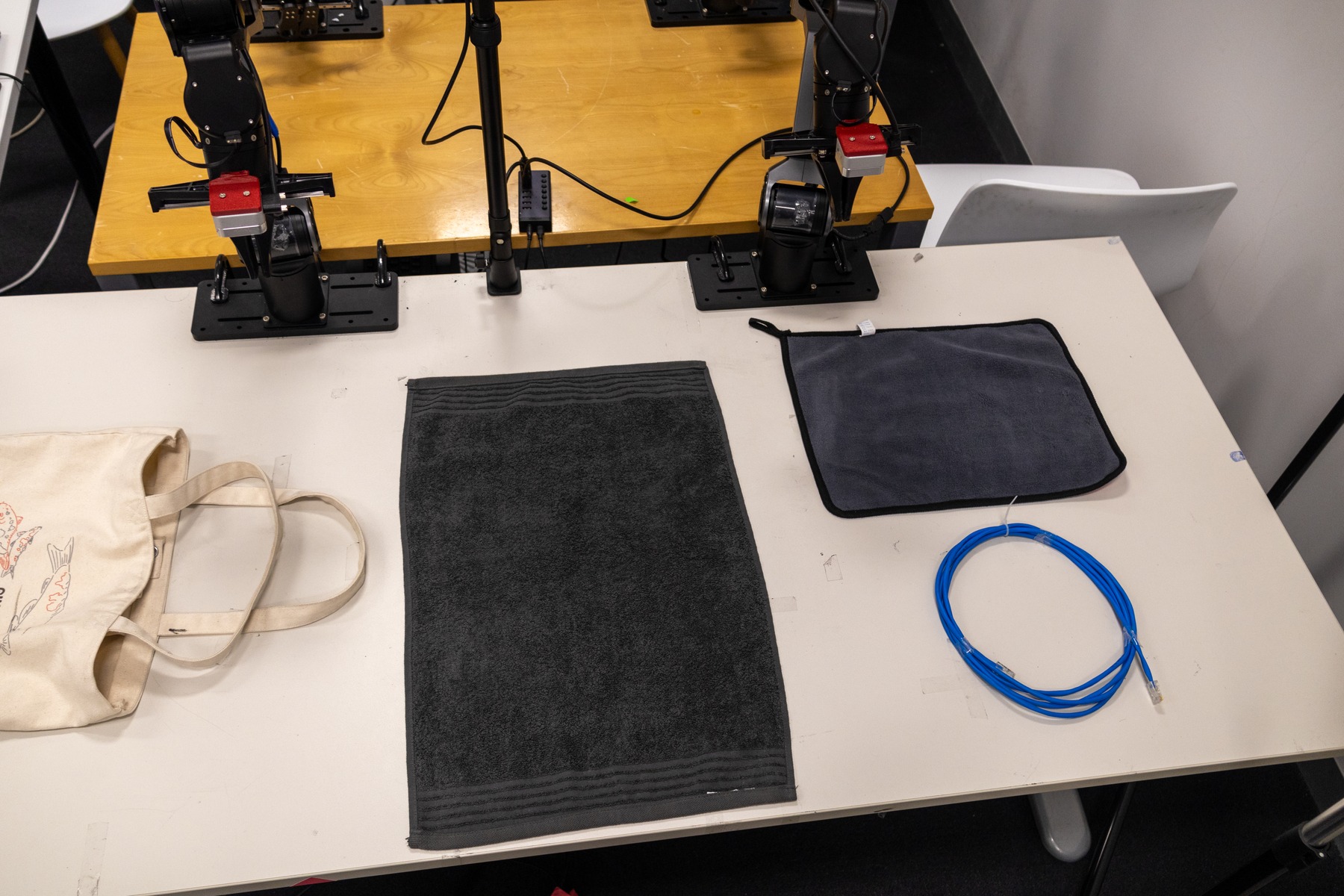}
\includegraphics[width=0.24\linewidth]{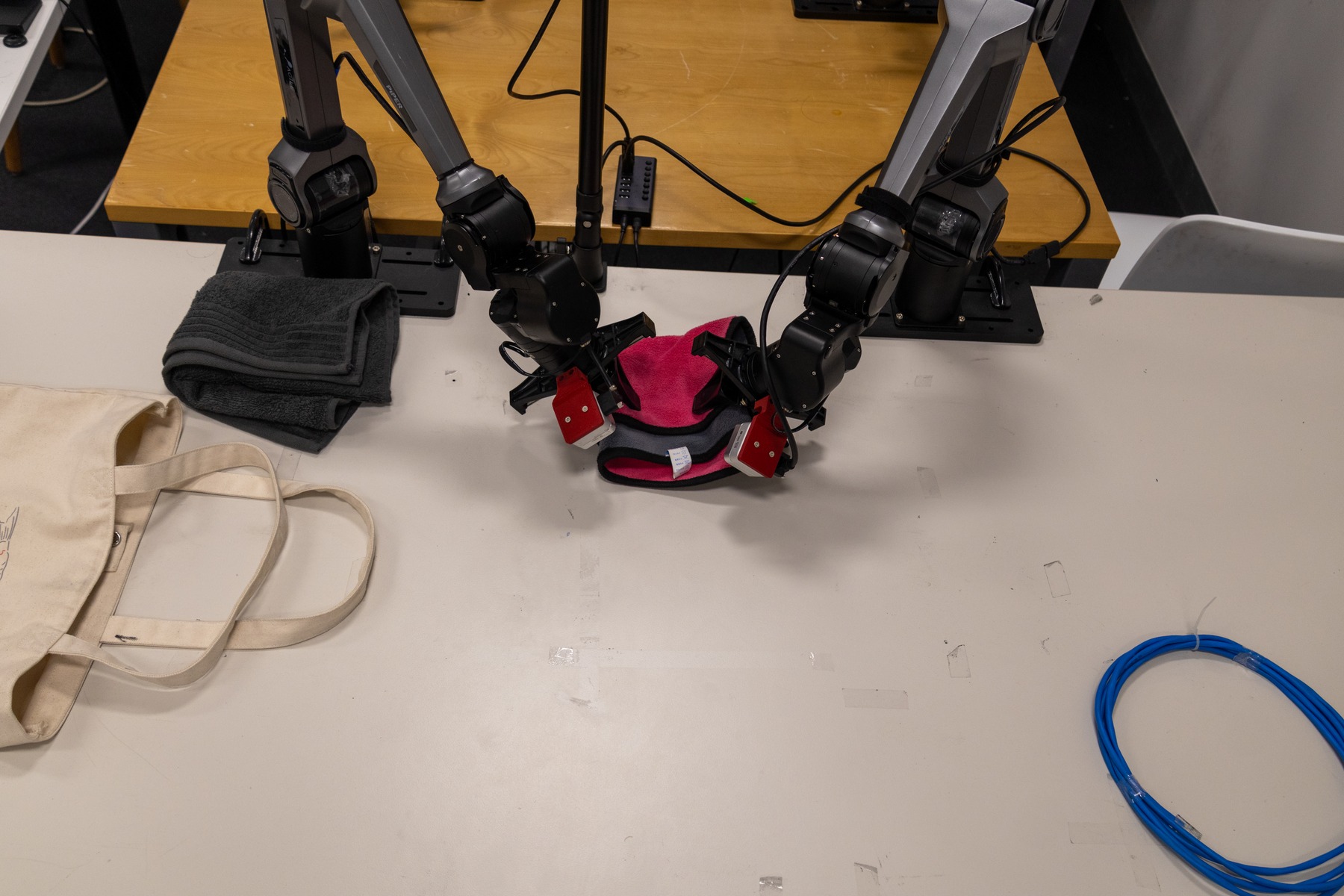}
\includegraphics[width=0.24\linewidth]{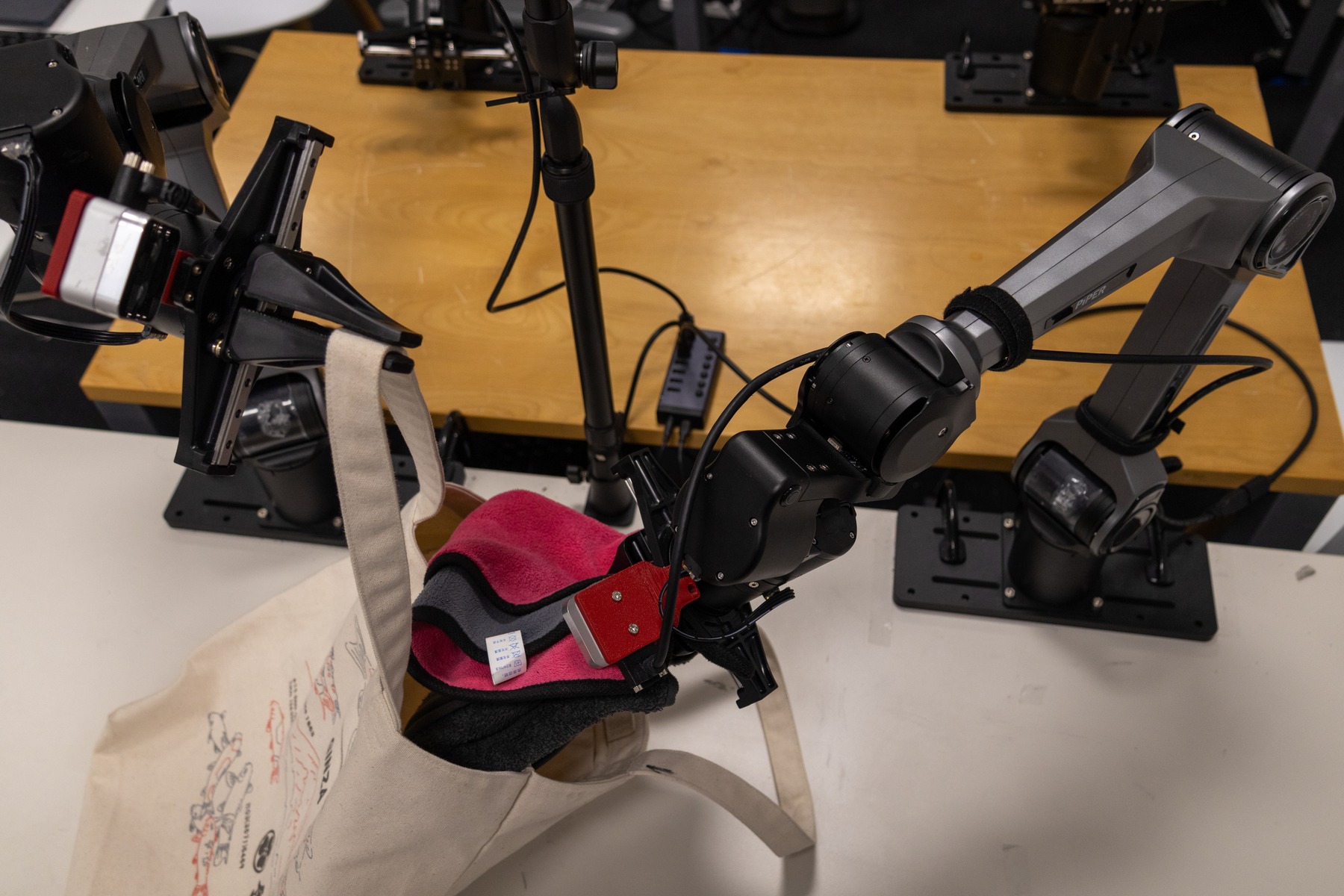}
\includegraphics[width=0.24\linewidth]{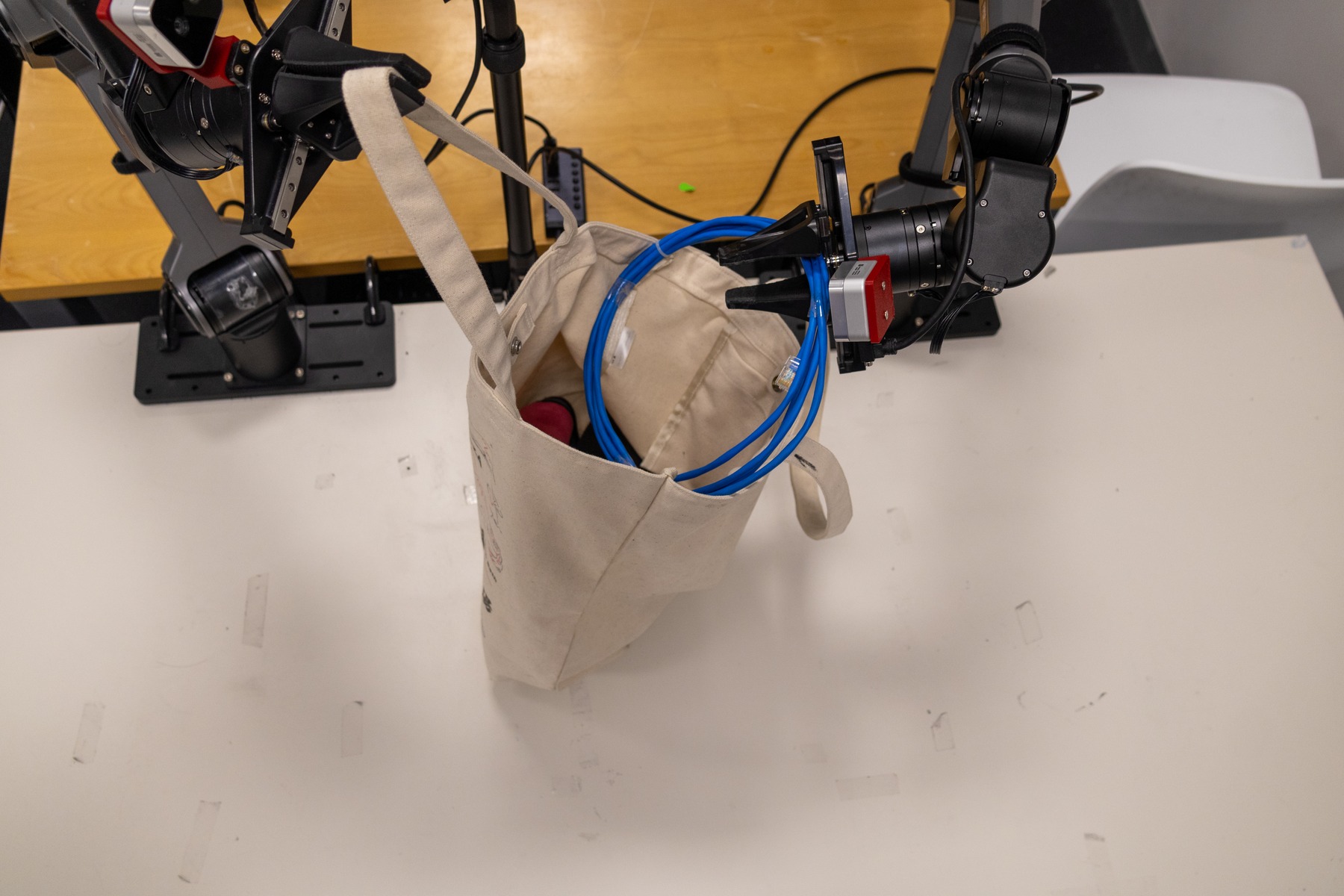}

\caption{Policies adapted with AutoIntervene handle deformable objects in the Towels-and-Cable Bagging task, including cloth, cables, and a tote bag.}\label{fig:long_horizon}
\end{figure*}

\begin{figure}[t]
    \centering
    \includegraphics[width=1.0\linewidth]{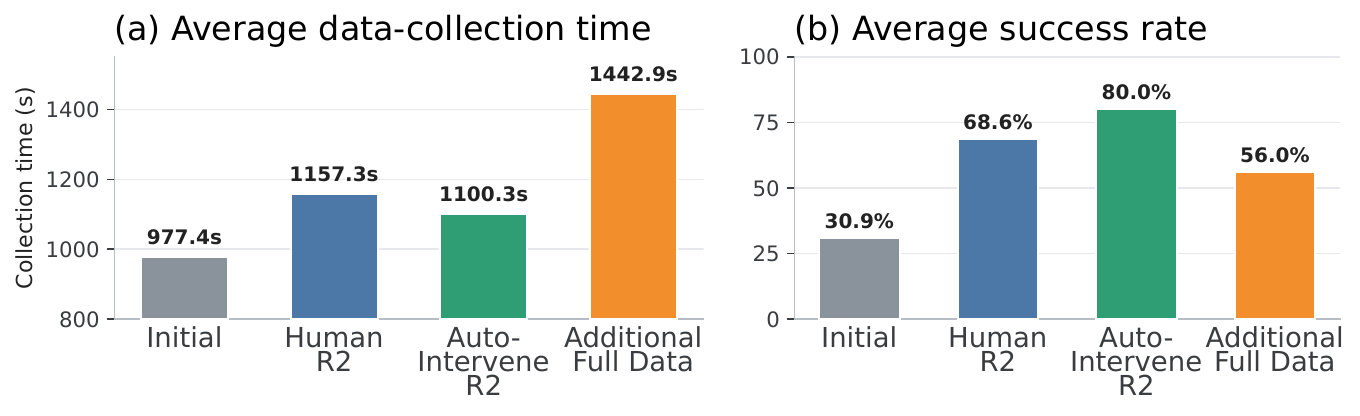}
    \caption{Across-task mean success and total recorded control-data time after R2, including the shared initial demonstrations.}
    \label{fig:BarCharts}
\end{figure}

We evaluate \emph{AutoIntervene} on nine real-world bimanual manipulation tasks, comprising a seven-task main benchmark and two longer-horizon extensions. The experiments are organized around four questions:
\textbf{Q1:} Does targeted intervention improve policy success relative to the initial policy and additional full demonstrations?
\textbf{Q2:} Does automatic switching achieve a better success--control-data trade-off than manual switching and collecting additional full demonstrations?
\textbf{Q3:} Can \emph{AutoIntervene} support iterative adaptation on long-horizon tasks and remain compatible with different action-generation heads?
\textbf{Q4:} How reliably does \emph{AutoIntervene} perform bidirectional handoff relative to prior monitors, and which components contribute to policy-to-operator and operator-to-policy switching?

\vspace{0.3em}\noindent\textbf{Experimental setup.}
All experiments used two AgileX PiPER-X 6-DoF arms with stock parallel grippers and three RGB cameras: one overhead and one on each wrist. Policy observations comprised a 28-dimensional bimanual state, including 14 joint-and-gripper values and their corresponding 14 measured-torque values, while actions were 14-dimensional joint-and-gripper commands. For the Diffusion Policy and Flow Matching variants, we retained the same robot, observation, and deployment interfaces while replacing only the action-generation head. Our fixed-base setup omitted the pedal and mobile-base channels of TriPilot-FF. We instantiated the policy's visual encoder with DINOv3 ConvNeXt-Base~\cite{simeoni2025dinov3}. We used two action groups corresponding to the left and right arms.

\begin{figure}[t]
\centering
\includegraphics[width=0.33\linewidth]{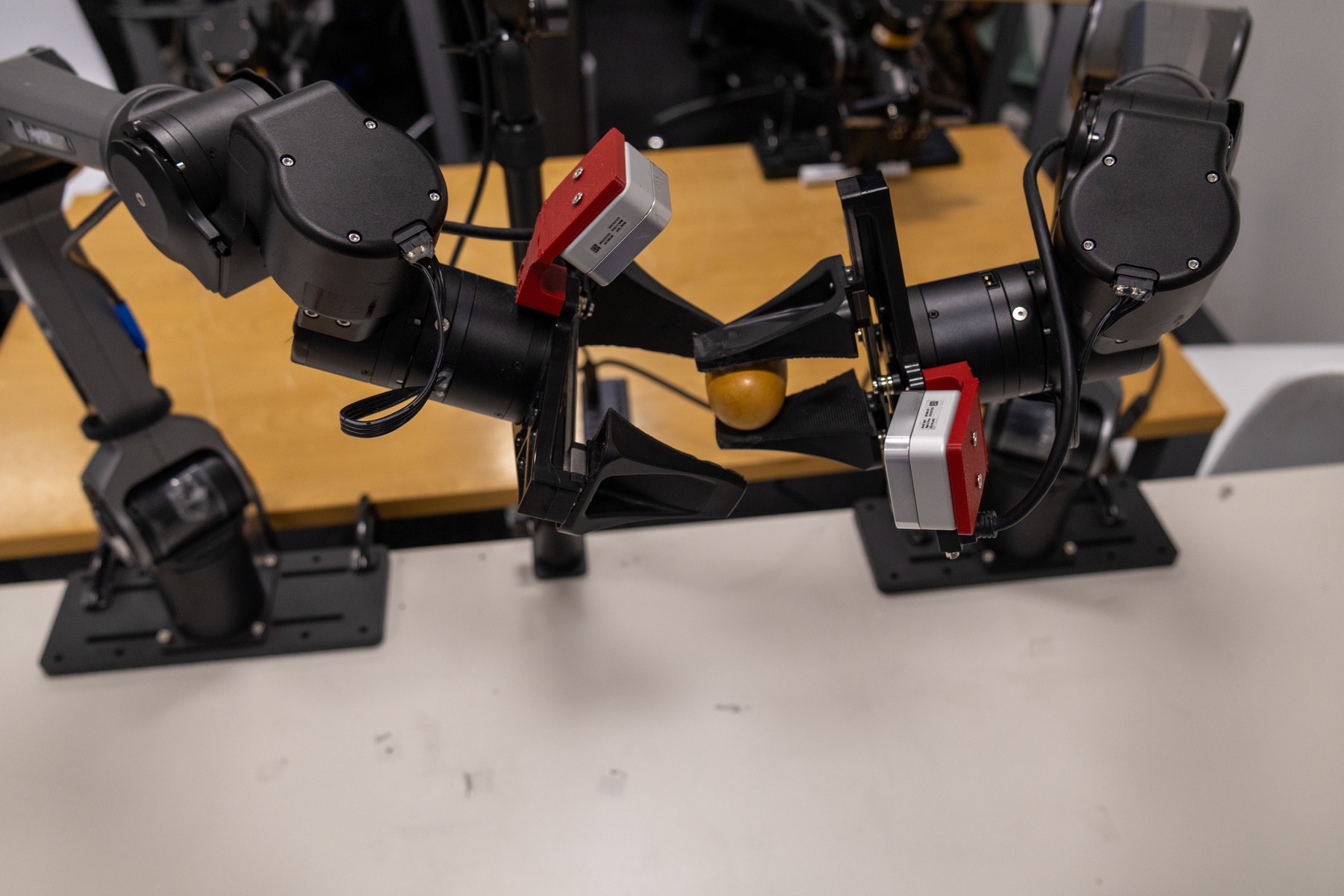}\includegraphics[width=0.33\linewidth]{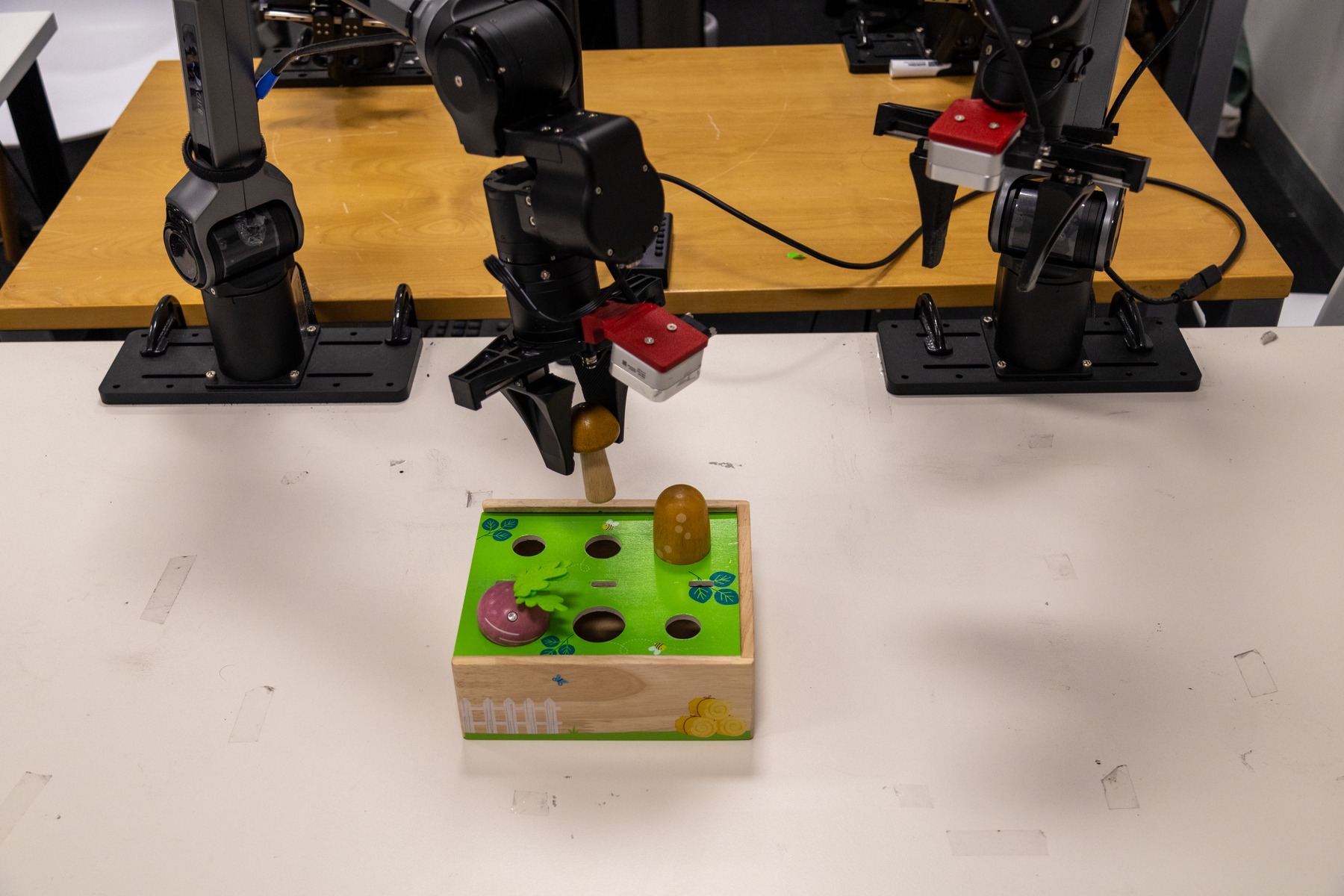}\includegraphics[width=0.33\linewidth]{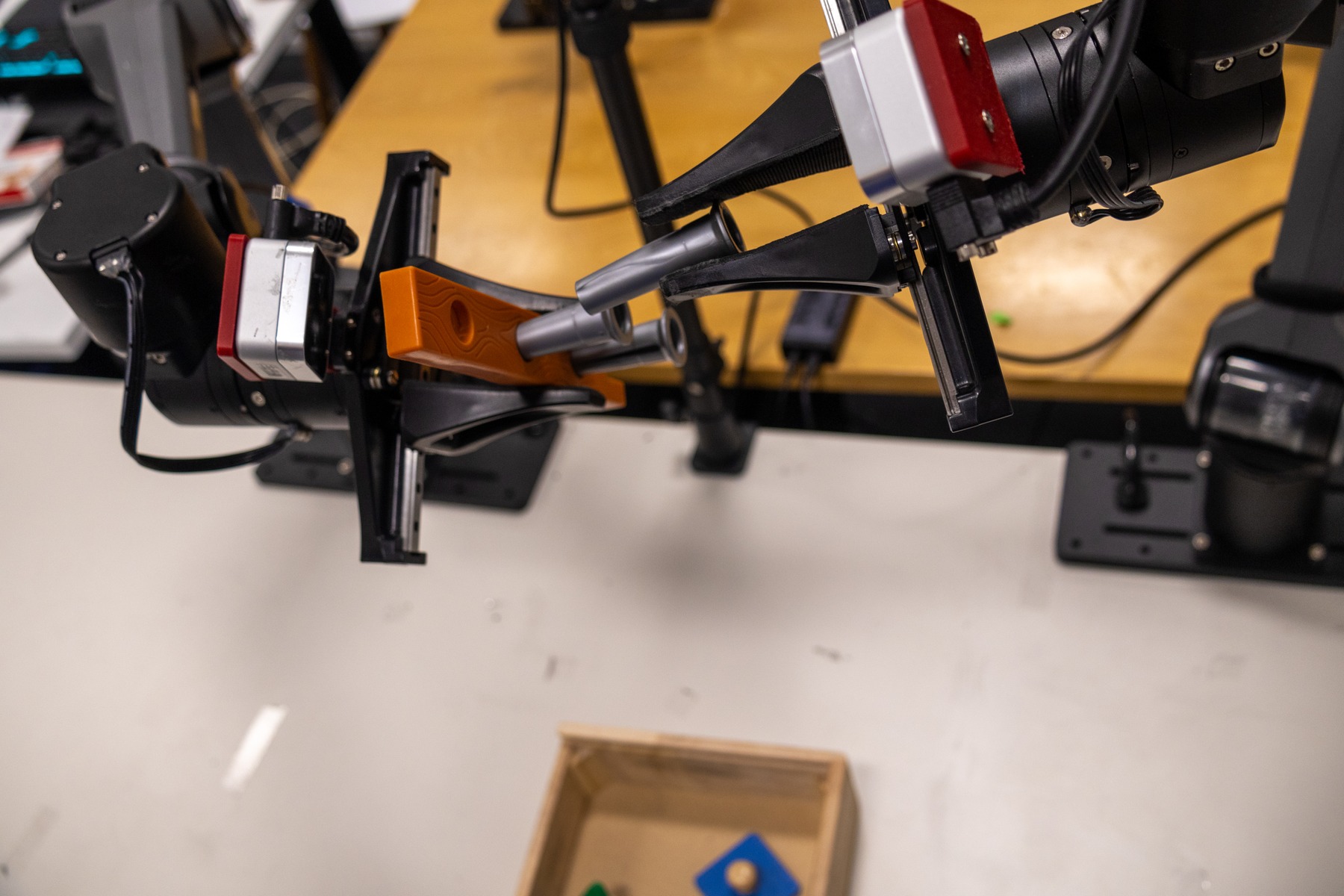}

\caption{Policies adapted with AutoIntervene perform precise manipulation in the Plant Sorting task, including insertion and handovers.}\label{fig:examples_exp}
\end{figure}

\vspace{0.3em}\noindent\textbf{Data and evaluation protocol.}
For each task, 30 of the 36 initial trajectories formed $\mathcal D^{(0)}$ and the remaining six formed the fixed $\mathcal D_{\mathrm{cal}}$. Subsequent intervention trajectories augmented both the training set and visual-action memory but not the fixed calibration set. For both Human and AutoIntervene, each adaptation round retained intervention trajectories from five successful deployment rollouts per task. Unless stated otherwise, every task-method success rate was measured over 25 unassisted physical rollouts. Success required completing every subtask in Table~\ref{tab:appendix_task_definitions}. The main benchmark comprised seven bimanual manipulation tasks, while two longer-horizon extensions evaluated iterative adaptation. Table~\ref{tab:appendix_task_definitions} defines all nine tasks. \Cref{fig:long_horizon,fig:examples_exp} show representative executions.

\vspace{0.3em}\noindent\textbf{Implementation details.}
Policies used $256\!\times\!256$ images and an action-chunk horizon of $H=100$ future actions. All policy variants used the same encoder--decoder backbone, consisting of one encoder layer and four decoder layers, while ACT, Diffusion Policy, and Flow Matching differed only in their action-generation mechanisms. For adaptation, accumulated pre-round training data and newly collected round-$k$ intervention data were sampled in a $2{:}1$ ratio ($\lambda_{\mathrm{mix}}=2/3$). The monitor used $J=K=16$, $B=H_r=40$, $M=3$, and $W=5$. After a 1\,s warm-up, under policy control, policy actions were executed at 30\,Hz, whereas AutoIntervene evaluation was performed at 5\,Hz. The phase-local retrieval windows were therefore updated every $U_{\mathrm{win}}=6$ executed policy actions. Under operator control, teleoperation commands were executed at 200\,Hz, whereas AutoIntervene evaluation was performed at 30\,Hz. Both directions required two consecutive decisions ($L_{\mathrm{pol}}=L_{\mathrm{op}}=2$). Each round recomputed its thresholds with $(\alpha_s^{\mathrm{pol}},\alpha_r^{\mathrm{pol}})=(0.05,0.05)$ and $(\alpha_s^{\mathrm{op}},\alpha_r^{\mathrm{op}})=(0.30,0.30)$.

\vspace{0.3em}\noindent\textbf{Baseline monitor settings.}
For the controlled comparison, we implemented deployment-time handoff monitors adapted from LazyDAgger~\cite{hoque2021lazydaggerreducingcontextswitching} and RND-DAgger~\cite{bire2024efficientactiveimitationlearning} to the ACT deployment setting, while following their original switching criteria and reported threshold-setting procedures. For RND-DAgger, we followed its reported threshold rule, set the threshold to twice the mean training-set RND score, and tested $W_{\mathrm{rec}}\in\{5,30\}$.

\vspace{0.3em}\noindent\textbf{Compared methods.}
The main benchmark compares four settings. \textbf{Initial} trains only on the 30 original training demonstrations. \textbf{Additional Full Data} adds ten full expert trajectories per task from the nominal initial-state distribution. \textbf{Human} uses the same intervention-learning pipeline as our method, but an operator manually selects policy-to-operator and operator-to-policy switches. Its retained intervention segments are mixed with previous training data as separate intervention trajectories. \textbf{\textsc{AutoIntervene}} uses the proposed visual-action support evaluation to trigger both switching directions automatically. For both intervention-based methods, R1 and R2 denote successive adaptation rounds.

\vspace{0.3em}\noindent\textbf{Metrics.}
The primary metric is task success rate. To measure data efficiency, we also report recorded control-data time. Table~\ref{tab:main_exp} reports demonstration time for Initial and Additional Full Data and additional operator-control time for Human and \textsc{AutoIntervene}. For all methods, \Cref{fig:BarCharts} reports total time by adding the shared initial demonstration time to the intervention methods.

\subsection{Targeted Intervention Improves Policy Adaptation (Q1)}
\label{sec:main_results}

Across the seven-task benchmark in Table~\ref{tab:main_exp}, targeted intervention consistently improves policy adaptation: after two rounds, both Human and \textsc{AutoIntervene} outperform the Initial policy and Additional Full Data in aggregate. \textsc{AutoIntervene} delivers the strongest improvement, increasing the across-task mean success rate by 49.1 percentage points, with gains across all seven tasks. These results support \textsc{AutoIntervene} as an effective approach for collecting targeted corrective supervision from learner-induced failure states.

\subsection{Automatic Switching Enables More Efficient Policy Adaptation (Q2)}
\label{sec:data_efficiency}

As shown in Table~\ref{tab:main_exp} and \Cref{fig:BarCharts}, \textsc{AutoIntervene} achieves higher average success than both Human and Additional Full Data, while using approximately $74\%$ less additional recorded control-data time than Additional Full Data. Relative to manual switching, \textsc{AutoIntervene} returns control to the policy once the recovered state and policy action return to demonstrated support, keeping each retained intervention segment focused on the failure and recovery. A manually selected operator-to-policy switch depends on operator judgment and button timing, so the segment can extend beyond recovery into states that the policy already handles. Training on this extra tail dilutes the corrective update by adding supervision in already-supported regions. Additional Full Data is also less targeted because it spends collection time on complete nominal trajectories rather than learner-induced failure states.

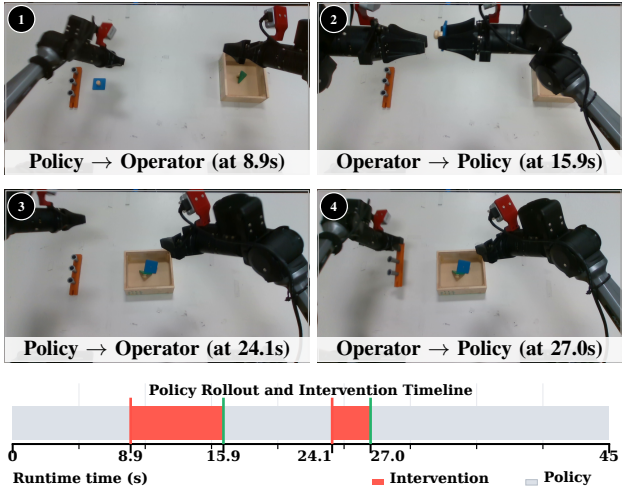
\begin{figure}[t]
    \centering

    \newlength{\RiskPanelW}
    \newlength{\RiskGap}
    \newlength{\RiskGridW}
    \setlength{\RiskPanelW}{0.47\columnwidth}
    \setlength{\RiskGap}{3pt}
    \setlength{\RiskGridW}{\dimexpr 2\RiskPanelW + \RiskGap\relax}

    \begin{minipage}{\RiskGridW}
    \centering

    \tikzset{
        riskpanelnode/.style={
            inner sep=0pt,
            outer sep=0pt
        },
        riskpanelbox/.style={
            draw=black!35,
            line width=0.45pt
        },
        riskbadge/.style={
            circle,
            fill=black,
            draw=white,
            line width=0.45pt,
            text=white,
            font=\bfseries\tiny,
            minimum size=0.36cm,
            inner sep=0pt
        },
        riskframetag/.style={
            fill=white,
            fill opacity=0.88,
            text opacity=1,
            font=\bfseries\footnotesize,
            align=center,
            text width=\RiskPanelW,
            inner xsep=0pt,
            inner ysep=1.0pt
        }
    }

    \newcommand{\RiskSeqPanel}[3]{\begin{tikzpicture}
            \node[riskpanelnode] (im)
                {\includegraphics[width=\RiskPanelW]{#2}};

            \draw[riskpanelbox]
                ([xshift=0.35pt,yshift=0.35pt]im.south west)
                rectangle
                ([xshift=-0.35pt,yshift=-0.35pt]im.north east);

            \node[riskbadge]
                at ([xshift=0.22cm,yshift=-0.22cm]im.north west)
                {#1};

            \node[riskframetag,above=0pt of im.south]
                {#3};
        \end{tikzpicture}}

    \includegraphics[width=\RiskGridW]{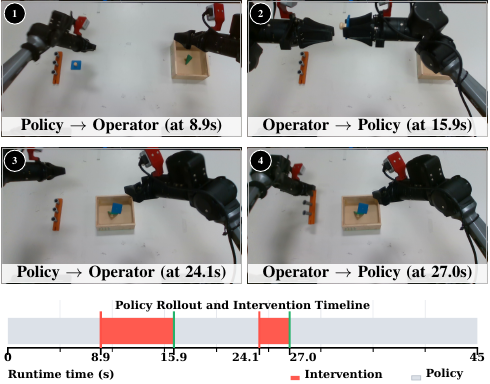}

    \end{minipage}

    \caption{One \emph{AutoIntervene} rollout on Peg Disassembly containing two policy--operator--policy intervention cycles. Top: the four control switches. Bottom: the corresponding timeline, with policy control in grey and operator control in red.}
    \label{fig:intervention_timeline_example}
\end{figure}

\Cref{fig:intervention_timeline_example} shows two policy--operator--policy cycles within a single Peg Disassembly trajectory, illustrating that \emph{AutoIntervene} can collect multiple intervention segments from one rollout.

\subsection{Iterative Adaptation for Longer-Horizon Tasks (Q3)}
\label{sec:iterative_refinement}

\begin{table}[t]
\centering
\caption{
Success (\%) over three AutoIntervene adaptation rounds on two long-horizon tasks.
}
\label{tab:intervention_refinement}
\small
\setlength{\tabcolsep}{8pt}

\begin{adjustbox}{width=0.9\linewidth}
\begin{tabular}{l|rr}
\toprule
Training Stage
& \shortstack{Two-Towel\\Box Packing}
& \shortstack{Towels-and-Cable\\Bagging}\\
\midrule
Initial Policy  & 28\% & 8\%  \\
AutoIntervene R1  & 44\% & 20\% \\
AutoIntervene R2  & 64\% & 28\% \\
AutoIntervene R3  & \textbf{88\%} & \textbf{48\%} \\
Additional Full Data & 52\% & 28\% \\
\bottomrule
\end{tabular}
\end{adjustbox}

\end{table}

\begin{table}[t]
\centering
\caption{
Success (\%) over three AutoIntervene adaptation rounds on Two-Towel Box Packing with Diffusion Policy (DP), Flow Matching (FM), and ACT action heads.
}
\label{tab:intervention_refinement_dp_fm}
\small
\setlength{\tabcolsep}{8pt}
\begin{tabular}{l|rrr}
\toprule
Training Stage
& DP
& FM & ACT\\
\midrule
Initial Policy        & 32\% & 32\% & 28\%\\
AutoIntervene R1      & 40\% & 36\% & 44\%\\
AutoIntervene R2      & 56\% & 48\% & 64\%\\
AutoIntervene R3      & \textbf{92\%} & \textbf{80\%} & \textbf{88\%}\\
Additional Full Data  & 44\% & 40\% & 52\%\\
\bottomrule
\end{tabular}
\end{table}

To evaluate \emph{AutoIntervene} on longer task executions, we perform three adaptation rounds on two longer-horizon manipulation tasks. As shown in Table~\ref{tab:intervention_refinement}, success improves after each round on both tasks, and the final policies outperform those trained with Additional Full Data. These results show that \emph{AutoIntervene} is not limited to correcting a single local failure: by collecting intervention data from the problems encountered during each deployment and using it to update the policy, it continues improving performance over longer, more complex task executions.

\subsection{Adaptation across ACT, Diffusion, and Flow-Matching Heads (Q3)}
\label{sec:action_heads}

To test whether \emph{AutoIntervene} transfers beyond ACT, we replace the ACT action head with Diffusion Policy and Flow Matching heads while keeping the shared backbone, robot interfaces, and monitor unchanged. As shown in Table~\ref{tab:intervention_refinement_dp_fm}, all three action heads improve over successive adaptation rounds, and their R3 policies outperform both the Initial policies and Additional Full Data. \emph{AutoIntervene} therefore transfers across different action-generation mechanisms without head-specific modification.

\subsection{AutoIntervene Outperforms Prior Handoff Monitors through Complementary Components (Q4)}
\label{sec:ablations}

\vspace{0.3em}\textbf{Protocol and metrics.}
Handoff is evaluated during lid opening in Lidded Box Packing using the same 20,000-step policy checkpoint, physical configuration, and pre-specified monitor parameters. Each method is tested in 10 rollouts after a fixed 5\,cm box translation and 10 nominal rollouts, with completion, 60\,s, or five handoff cycles terminating a rollout. The perturbed condition tests whether the monitor transfers control to the operator after a policy failure and returns it to the policy after operator recovery. The nominal condition tests whether a successful autonomous execution remains uninterrupted when no intervention is needed. A valid \textbf{cut-in} occurs between displacement and 3\,s after the resulting failed grasp. A valid \textbf{cut-out} is the operator-to-policy switch after the operator restores the box and guides the gripper to the lid. For the four transition metrics on perturbed rollouts, let $V_d$, $M_d$, and $E_d$ denote the numbers of valid transitions, missed opportunities, and extra transitions for $d\in\{\text{cut-in},\text{cut-out}\}$. We compute
\begin{equation}
\operatorname{Recall}_d=\frac{V_d}{V_d+M_d},\qquad
\operatorname{Precision}_d=\frac{V_d}{V_d+E_d}.
\label{eq:handoff_metrics}
\end{equation}
For the metrics in \eqref{eq:handoff_metrics}, a zero denominator is reported as N/A. A valid cut-out must follow a valid cut-in. Consequently, cut-out recall is conditioned on valid cut-ins, whereas cut-out precision considers all attempted operator-to-policy switches. The false-trigger rate is the fraction of nominal rollouts containing at least one unnecessary cut-in, with each rollout counted at most once regardless of repeated triggers.

The comparison includes LazyDAgger~\cite{hoque2021lazydaggerreducingcontextswitching} and RND-DAgger~\cite{bire2024efficientactiveimitationlearning}, with the two RND-DAgger settings differing only in recovery persistence $W_{\mathrm{rec}}$. The ablations preserve retrieval and neighbour selection, removing only visual support or action risk from the acceptance criteria.

\begin{table}[t]
\centering
\caption{Controlled handoff comparison on Lidded Box Packing using 10 perturbed and 10 nominal rollouts per method, capped at five complete handoff cycles. Values are rates. N/A indicates an undefined denominator.}
\label{tab:handoff_ablation}
\scriptsize
\setlength{\tabcolsep}{1.5pt}
\renewcommand{\arraystretch}{1.05}
\begin{adjustbox}{max width=\columnwidth}
\begin{tabular}{@{}lccccc@{}}
\toprule
& \multicolumn{4}{c}{Perturbed} & \multicolumn{1}{c}{Nominal} \\
\cmidrule(lr){2-5}\cmidrule(lr){6-6}
Method
& \shortstack{Cut-in\\Recall $\uparrow$}
& \shortstack{Cut-in\\Prec. $\uparrow$}
& \shortstack{Cut-out\\Recall $\uparrow$}
& \shortstack{Cut-out\\Prec. $\uparrow$}
& \shortstack{False-trigger\\Rate $\downarrow$} \\
\midrule
AutoIntervene & \textbf{1.00} & \textbf{1.00} & \textbf{1.00} & \textbf{1.00} & \textbf{0.00} \\
\midrule
\multicolumn{6}{@{}l}{\emph{Prior handoff monitors}} \\
LazyDAgger~\cite{hoque2021lazydaggerreducingcontextswitching} & 0.40 & \textbf{1.00} & 0.00 & N/A & 0.80 \\
RND-DAgger ($W_{\mathrm{rec}}=5$)~\cite{bire2024efficientactiveimitationlearning} & 0.80 & 0.16 & 0.00 & 0.00 & 0.90 \\
RND-DAgger ($W_{\mathrm{rec}}=30$)~\cite{bire2024efficientactiveimitationlearning} & 0.80 & 0.16 & 0.75 & 0.12 & 0.90 \\
\midrule
\multicolumn{6}{@{}l}{\emph{Ablations of visual support and action risk}} \\
w/o visual support & 0.00 & N/A & N/A & N/A & \textbf{0.00} \\
w/o action risk & 0.90 & \textbf{1.00} & 0.56 & 0.56 & \textbf{0.00} \\
\bottomrule
\end{tabular}
\end{adjustbox}
\end{table}

\vspace{0.3em}\textbf{Comparison with prior monitors.}
As shown in Table~\ref{tab:handoff_ablation}, reliable bidirectional handoff without unnecessary policy-to-operator switches during nominal execution is achieved only by AutoIntervene. LazyDAgger misses failures and retains operator control because policy--operator disagreement remains large during recovery. RND-DAgger uses one novelty threshold for both directions. After an operator-to-policy switch, a score increase can cross the same threshold and immediately return control to the operator. Increasing recovery persistence only delays the return to policy control, whereas AutoIntervene uses separately calibrated, mode-specific switching criteria to implement hysteretic switching and suppress immediate reversal.

\vspace{0.3em}\textbf{Ablations of visual support and action risk.}
As shown in Table~\ref{tab:handoff_ablation}, without visual support, \emph{AutoIntervene} misses the workspace displacement because action risk can remain low despite weak visual correspondence. Without action risk, most policy-to-operator switches remain, but the return to policy control may be less reliable when visual correspondence recovers before the proposal matches successful reference actions. Thus, visual support detects when intervention is needed and action risk prevents a premature operator-to-policy switch. Both are required for reliable bidirectional handoff.

\noindent\textbf{Ablation of policy-side retrieval-window updates.}
Beyond the score components, we evaluate policy-side retrieval-window updates by lowering the target object while preserving a similar RGB appearance, causing the nominal grasp to fail. Across 10 rollouts per setting, \emph{AutoIntervene} with retrieval-window updates achieves a cut-in recall of 1.00. During repeated-grasp failures, the updates advance the phase-local windows until they all reach their final valid chunks, incrementing the policy-side rejection counter until control transfers to the operator. Without these updates, the policy can repeat failed grasps without transferring control, yielding a cut-in recall of 0.30.

\section{Conclusion}
AutoIntervene combines phase-local visual-action monitoring, bidirectional handoff, and intervention-data aggregation. Across nine real-world bimanual tasks, it improved policies over successive rounds with less additional control data than full demonstrations, achieved higher mean success than manual switching, and required less operator-control time on average. Together with separately calibrated criteria for both switching directions, these components focus operator control on unsupported periods and resume autonomy once the policy proposal returns to demonstrated support. Future work will extend online calibration across broader tasks, perturbations, and operators.

\bibliographystyle{IEEEtran}
\bibliography{references}

\end{document}